\pdfoutput=1
\documentclass[sigconf]{acmart}

\usepackage{multirow}
\usepackage{balance}
\AtBeginDocument{%
  \providecommand\BibTeX{{%
    \normalfont B\kern-0.5em{\scshape i\kern-0.25em b}\kern-0.8em\TeX}}}

\copyrightyear{2021}
\acmYear{2021}
\setcopyright{acmcopyright}\acmConference[MM '21]{Proceedings of the 29th ACM International Conference on Multimedia}{October 20--24, 2021}{Virtual Event, China}
\acmBooktitle{Proceedings of the 29th ACM International Conference on Multimedia (MM '21), October 20--24, 2021, Virtual Event, China}
\acmPrice{15.00}
\acmDOI{10.1145/3474085.3475274}
\acmISBN{978-1-4503-8651-7/21/10}

\acmSubmissionID{mfp0576}

\begin{document}
\fancyhead{}

\title{
Attribute-specific Control Units in StyleGAN for Fine-grained Image Manipulation
}
%
%
%\author{Rui Wang}
%\affiliation{
%    \department[0]{Key Laboratory of Image Processing and Intelligent Control}
%    \department[1]{School of Artificial Intelligence and Automation}
%    \institution{Huazhong University of Science and Technology}
%  \city{Wuhan}
%  \country{China}
%}
%\email{autowangrui@hust.edu.cn}
%
%
%\author{Jian Chen}
%\author{Gang Yu}
%\affiliation{%
%  \institution{Tencent}
%  \city{Shanghai}
%  \country{China}
%}
%\email{urichen@tencent.com}
%\email{skicy@outlook.com}
%
%
%\author{Li Sun}
%\affiliation{%
%  \institution{East China Normal University}
%  \city{Shanghai}
%  \country{China}
%}
%\email{sunli@ee.ecnu.edu.cn}
%
%
%\author{Changqian Yu}
%\author{Changxin Gao}
%\authornote{Corresponding author.}
%\author{Nong Sang}
%\affiliation{
%    \department[0]{Key Laboratory of Image Processing and Intelligent Control}
%    \department[1]{School of Artificial Intelligence and Automation}
%    \institution{Huazhong University of Science and Technology}
%  \city{Wuhan}
%  \country{China}
%}
%\email{{changqian_yu, cgao, nsang}@hust.edu.cn}

 \author{Rui Wang$^1$, Jian Chen$^2$, Gang Yu$^2$, Li Sun$^3$, Changqian Yu$^1$, Changxin Gao$^{1}\dagger$, Nong Sang$^1$}
 \affiliation{%
   \institution{
$^1$Key Laboratory of Image Processing and Intelligent Control, School of Artificial Intelligence and Automation, Huazhong University of Science and Technology, Wuhan \\
$^2$Tencent, Shanghai
               $^3$East China Normal University, Shanghai \\
               }
\country{China}
   }
 \email{{autowangrui, changqian_yu, cgao, nsang}@hust.edu.cn, urichen@tencent.com, skicy@outlook.com, sunli@ee.ecnu.edu.cn}


\renewcommand{\shortauthors}{Wang, et al.}

\begin{abstract}

Image manipulation with StyleGAN has been an increasing concern in recent years.
Recent works have achieved tremendous success in analyzing several semantic latent spaces to edit the attributes of the generated images.
However, due to the limited semantic and spatial manipulation precision in these latent spaces, the existing endeavors are defeated in fine-grained StyleGAN image manipulation, i.e., local attribute translation.
To address this issue, we discover attribute-specific control units, which consist of multiple channels of feature maps and modulation styles. 
Specifically, we collaboratively manipulate the modulation style channels and feature maps in control units rather than individual ones to obtain the semantic and spatial disentangled controls. 
Furthermore, we propose a simple yet effective method to detect the attribute-specific control units. 
We move the modulation style along a specific sparse direction vector and replace the filter-wise styles used to compute the feature maps to manipulate these control units.
We evaluate our proposed method in various face attribute manipulation tasks. Extensive qualitative and quantitative results demonstrate that our proposed method performs favorably  against the state-of-the-art methods. The manipulation results of real images further show the effectiveness of our method. 

\end{abstract}

% \input{sections/00-abstract}



%
% This command processes the author and affiliation and title
% information and builds the first part of the formatted document.
\maketitle

\section{Introduction}

StyleGAN~\cite{karras2019style, karras2020analyzing}, as a state-of-the-art GAN, can synthesize diverse images with high quality.
The StyleGAN generator maps the random sampled noises to the intermediate latent codes and then further transforms them into hierarchical styles that modulate the hidden features.
By analyzing and identifying semantics in the latent representations of StyleGAN, we can reuse existing pre-trained StyleGAN to manipulate the 
synthesized images, which has a wide range of applications.

\begin{figure}[t]
\begin{center}
\includegraphics[width=1.0\linewidth]{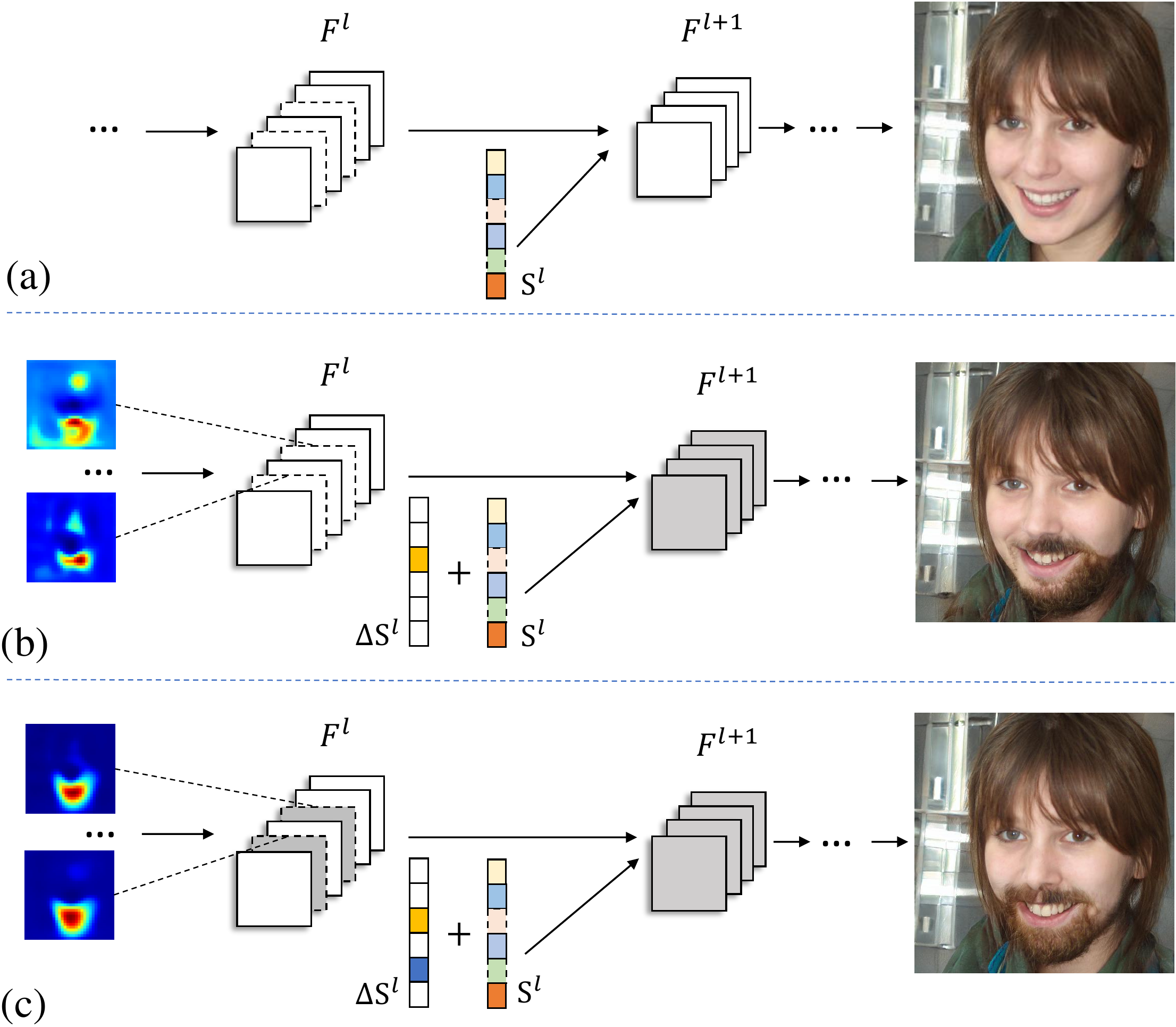}
\end{center}
\caption{
Both modulation styles and feature maps in control units need to be manipulated to add a full beard.
(a) The original StyleGAN generation process. 
(b) Manipulate a single channel of the $l^{th}$ layer's modulation styles $S^l$.
(c) Manipulate a few channels of $S^{l}$ and their corresponding modulated feature maps collaboratively.
The control units are indicated by the dashed border.
Gray feature maps indicate that they have been modified.
}\label{fig:fig1}
\end{figure}
%Due to the advantages of StyleGAN, in recent years, attribute manipulation with StyleGAN has achieved great success by constructing some semantic latent spaces~\cite{} and discovering the controls in these spaces~\cite{}. 
The key problem in attribute manipulation with StyleGAN is how to obtain a fine-grained controls, i.e., semantic and spatial disentangled controls. 
A well-trained StyleGAN inherently encodes numerous semantic information in the latent space and hidden features\cite{radford2015unsupervised,upchurch2017deep, bau2019gandissect,shen2020interpreting,shen2020interfacegan,yang2021semantic,xu2021generative}. %, despite the absence of explicit supervision during training.
GAN Dissection~\cite{bau2019gandissect} found the causal feature maps are specialized to synthesize specific visual concepts in generated images.
Some previous works~\cite{goetschalckx2019ganalyze,jahanian2019steerability,harkonen2020ganspace,wang2021hijack} find that rich semantic information is encoded in the latent space of StyleGAN, e.g., $\mathcal{Z}$ or $\mathcal{W}$ space and various semantic manipulations can be achieved by moving the latent code along the direction in the latent space.
But modifications to the latent code in the $\mathcal{Z}$ or $\mathcal{W}$ space are spatial entangled. 
%lack explicit control over the modified region. 
This means, while the target attribute of the generated images is successfully manipulated using these methods, irrelevant regions and attributes in the image are usually also changed. 
To address this issue, very recent works~\cite{wu2020stylespace, liu2020style} further introduce the $\mathcal{S}$ space spanned by the modulation styles.
% %
% StyleSpace~\cite{wu2020stylespace} finds the feature map's modulation styles at different scales are rich in semantic information and more disentangled than other intermediate latent representations.
% %
StyleSpace~\cite{wu2020stylespace} measures and compares the disentanglement and completeness of these spaces using the metrics proposed by ~\cite{eastwood2018framework}, and finds that the feature map's modulation styles at different scales are rich in semantic information.
Spatial disentangled manipulation can be achieved by modifying only a single channel of the manipulation style.
%
%We believe that this is because adjusting the modulation parameters of a single channel brings about a change in the following feature maps consistent with this channel of feature's spatial distribution.
%As shown in Figure~\ref{fig:fig1}(b), editing the corresponding single-channel style control is able to add the attribute ``goatee'', and only the region around the mouth obviously changes. %is mainly  Although We can see that the single-channel style control, that is modulation style~\cite{wu2020stylespace} or feature map~\cite{bau2019gandissect} of a channel, bring consistent attribute manipulations on the diverse generated images.
%
However, the results manipulated by these single-channel controls still suffer from the \textit{insufficient change issue}.
As shown in Figure~\ref{fig:fig1}(b), although we successfully added the attribute ``goatee'', the shape of the goatee is not complete enough.
We observe that the top activated regions of the corresponding feature map are incomplete according to the region of the attribute.

Inspired by the above observations, we infer that the semantics of a specific region in the image is encoded both in the channels of intermediate feature maps and the corresponding modulation styles. 
And modifying high activation parts of the feature maps to align with the attribute related region is able to ``completely'' edit the attribute region. 
As shown in Figure~\ref{fig:fig1}(c), after modifying the feature maps, we obtain a full goatee.
In this paper, we discover \textit{attribute-specific control units}, which consist a few channels of the modulation styles and feature maps.
We collaboratively manipulate the modulation styles and feature maps rather than individual ones, to obtain the fine-grained controls.

The image change caused by the modification to the modulation style for a specific intermediate layer in StyleGAN has a similar spatial structure to the activation, as demonstrated in Figure~\ref{fig:fig1}. 
We divided each channel of the intermediate features into different region-specific groups based on the spatial location of the top activated region of the feature map with a simple yet effective gradient-based strategy. 
We then restrict manipulations to the channel group corresponding to the target semantic region to prevent the change of unrelated regions of generated images. 
The resulted sparse direction is then used to manipulate the modulation channels in control units.
% %
% To manipulate the modulation style channels in control units, we construct a spare direction with the dominated channels within the channel group corresponding to the target semantic region of the modulation style direction. 
%
We also replace the modulation style used to compute the crucial feature maps with a style vector obtained by optimization. 
The activated region of feature maps in control units aligns precisely and consistently with the target region with the help of this operation. 
As a result, our proposed method can bring the disentangled and effective fine-grained controls on local translations.
Our main contributions could be summarized as follows:

\begin{itemize}
    \item We demonstrate that the specific semantic region's attribute is controlled by a few channels of intermediate feature and its corresponding modulation styles, which are represented as control units.
    
    \item We propose a simple yet effective method to detect the attribute-specific units. We move the modulation style along a specific sparse direction vector and replace the modulation style used to compute the crucial feature maps to manipulate these control units.
    
   \item We validate the effectiveness of our approach with the task of facial attribute editing. Extensive qualitative and quantitative results show that our method can achieve effective and spatial disentangled StyleGAN controls on local translations.
\end{itemize}

\section{Related Work}

% \subsection{Latent Representations of GAN}

% Visualization of specific modification
\begin{figure*}[t]
\begin{center}
\includegraphics[width=1.0\linewidth]{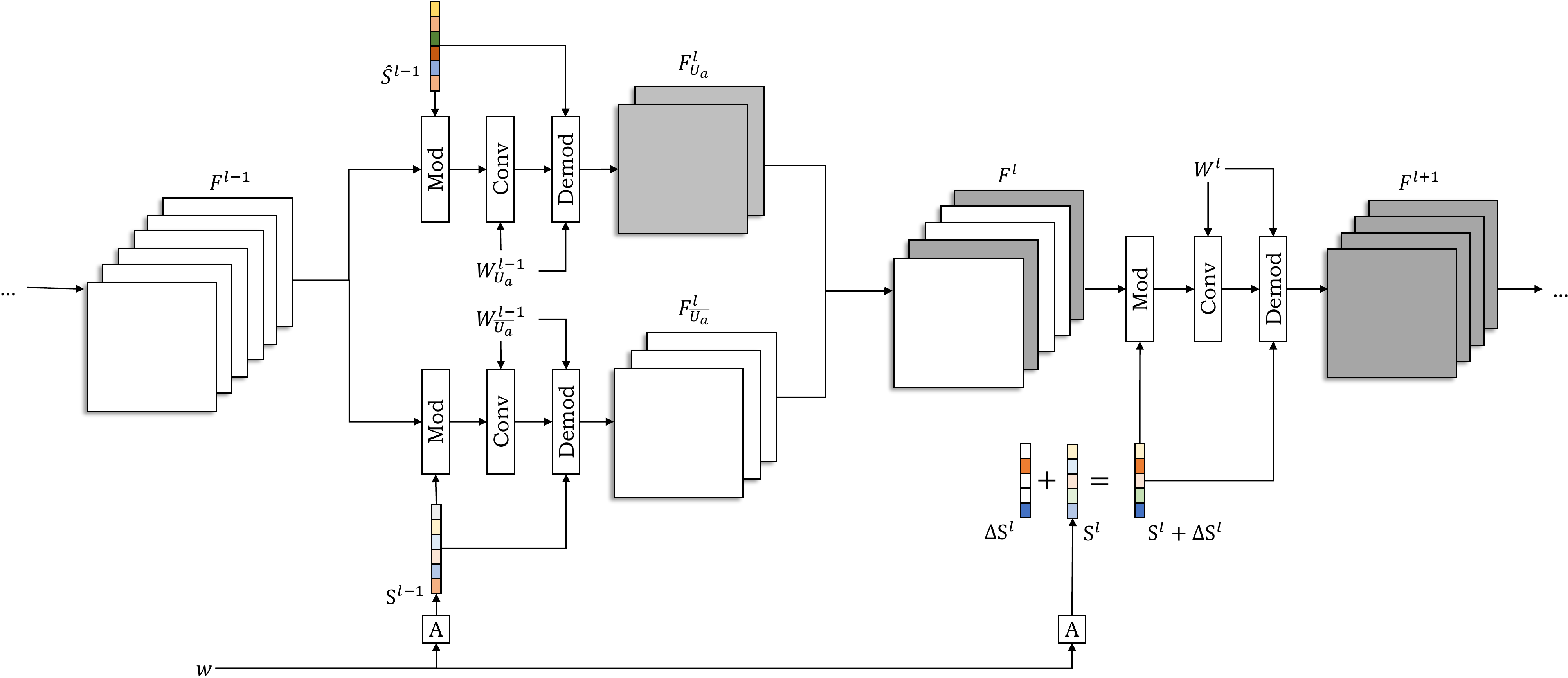}
\end{center}
\caption{
Visualization of a typical attribute manipulation pipeline.
We equivalently adjust the order of modulation, demodulation, and convolution operation for convenience.
Our modification contains a sparse direction vector $\Delta{S^l}$ and a modulation style vector $\hat{S}^{l-1}$ obtained by optimization.
The gray feature maps indicate that they have changed because of our modification.
}
\label{fig:method}
\end{figure*}

Manipulating the latent representation in StyleGAN can be used to edit the semantics of the synthesized images.
Real images also can be inverted into the latent spaces of StyleGAN by GAN Inversion methods~\cite{abdal2019image2stylegan,abdal2020image2stylegan++,richardson2020encoding,zhu2020indomain,tov2021designing}, enabling the further manipulations.
The approaches of semantic image editing with StyleGAN can be roughly divided into two groups, i.e., supervised approaches and unsupervised approaches.
The supervised approaches~\cite{goetschalckx2019ganalyze, jahanian2019steerability,shen2020interfacegan,liu2020style,hou2020guidedstyle,abdal2021styleflow} introduce pretrained classifiers to find the directions that alter the output of the classifiers.
For example, InterfaceGAN~\cite{abdal2019image2stylegan} trains linear support vector machines (SVMs) using the attribute annotations labeled by the off-the-shelf classifiers and finds hyperplanes in the latent space serving as the separation boundary.
StyleFlow~\cite{abdal2021styleflow} uses normalizing flow to establish a bi-directional mapping of noise and attributes to latent.
The unsupervised approaches~\cite{harkonen2020ganspace, voynov2020unsupervised, lu2020unsupervised, cherepkov2020navigating} aim to discover as many directions as possible using unsupervised techniques. 
GANSpace~\cite{harkonen2020ganspace} adopts PCA in latent space and identifies important latent directions.
\cite{shen2021closedform,spingarn2020gan} further show that meaningful directions can be computed in closed form directly from the generator's weights without any form of training or optimization.
Recent methods~\cite{collins2020editing,liu2020style,wu2020stylespace} reveal that the walks in $\mathcal{S}$ space facilitates the
spatial disentanglement in the spatial dimension.
\cite{collins2020editing} accomplished the local semantically-aware edits by transferring modulation style between source and target image.
StyleSpace~\cite{wu2020stylespace} shows that fine-grained controls can be obtained by only modifying the value of one single detected channel.
However, modifying only one channel may also affect multiple attributes of the image or only have a slight effect on the synthesized image's target attribute.

In this paper, we not only discover meaningful control directions that involve multiple style channels, but also simultaneously edit the modulated feature maps for fine-grained local attribute manipulation.

\section{Method}

In this paper, we investigate the fine-grained local attribute manipulation problem, which aims to edit one local attribute over the images synthesized by well-trained StyleGAN2, without affecting other attributes.
%aim to achieve fine-grained local control to the images synthesized by well-trained StyleGAN2. 
%
To this end, we propose to %reasonably 
manipulate the attribute-specific control units, which consider the feature maps of a few channels and the corresponding modulation style channels simultaneously, to synthesize certain attributes of the semantic region.
In this section, we first introduce the proposed attribute-specific control units, and then we describe how to detect and manipulate these control units, respectively.
%In Sec.~\ref{sec:why}, we first demonstrate why modifying only a few units enables fine-grained manipulations.
% %
% Then in Sec.~\ref{sec:movement} and Sec.~\ref{sec:replacement}, 
%Then in next sections, we describe how to detect and manipulate these crucial units, respectively.

\subsection{Attribute-Specific Control Units in StyleGAN2}\label{sec:why}

StyleGAN2~\cite{karras2020analyzing} involves multiple latent spaces in learning the mapping $G: \mathcal{Z}\to\mathcal{X}$. $\mathcal{X}$ stands for the image space.
The random noise vectors $z\in\mathcal{Z}$ are transformed into the intermediate latent space $\mathcal{W}$. 
Each $w\in\mathcal{W}$ is specialized to the hierarchical \textit{modulation styles} $S$ in $\mathcal{S}$ space~\cite{wu2020stylespace} by the learned affine transformations in each layer of StyleGAN2.
The intermediate features in the synthesis network of StyleGAN2 are channel-wisely modulated by $S\in\mathcal{S}$.
The hierarchical styles controlled semantics at different levels of the synthesized image.
As described in ~\cite{karras2019style,karras2020analyzing},
the different layers in StyleGAN control the different content of generated images.
Therefore, not all modulation styles for every layer need to be modified to manipulate a certain local attribute.

Suppose we aim to manipulate the attribute of a specific semantic region $r$ and the modulation styles after the $l^{th}$ layer do not affect the target attribute. 
Let us denote the ${l+1}^{th}$ layer's input feature as $F^{l+1}$, and using $\mathcal{F}^{l+1}$ to present the output
of the convolution layer in the $l^{th}$ layer.
Our goal is to precisely manipulating pixels of $F^{l+1}$ in the region $r$, while keeping the pixels in the other areas unchanged.

$F^{l+1}$ is calculated by the $l^{th}$ layer in the synthesis network of StyleGAN2 from the feature $F^{l}$ and the modulation style $S^l$.
When the modulation style $S^{l}$ walks along the direction vector $\Delta{S^l}$, the corresponding change in the convolution layer's output $\mathcal{F}^{l+1}$ is approximately in proportion to the modulated $F^{l+1}$ by
$\Delta{S}^{l}$.
Take the $k^{th}$ channel of $\mathcal{F}^{l+1}$ as an example:
\begin{equation}\label{eq:delta_F_2}
    \Delta{\mathcal{F}}^{l+1}_k \propto \sum_i^{n^{l}}\Delta{s_i^l}(W_{k,i}^l\star{F_i^l})
\end{equation}
where $\Delta{S}^l=[\Delta{s}_1^l, \cdots, \Delta{s}^l_{n^l}]^{T}$ has the same dimension as the number of channels $n^l$ of $F^l$, $\star$ denotes the convolution operation.
(Detail proof can be found in \textbf{supplementary material}).
The modulation, convolution and the non-linear components between $\mathcal{F}^{l+1}$ and $F^{l+1}$ are both spatially independent. 
Therefore, the spatial distribution of $\Delta{F}^{l+1}$ caused by the direction $\Delta{S^l}$ is highly related to the spatial distribution of each channel of $F^{l}$, which is consistently activated in a particular semantic region of generated images~\cite{simonyan2013deep,bau2017network,bau2019gandissect}.
Let us use $U_r$ to represent the set of indices of these feature channels that are consistently activated in target region $r$ across various generated images.
According to Eq.~\ref{eq:delta_F_2}, 
the change to the region $r$ in $F^{l}$ is mainly caused by the change of $S_{U_r}$.
Therefore, we can set the value of channels not in $U_r$ of $\Delta{S}^l$ as 0 to prevent the modification to unrelated regions.
We denote the set of indices of the dominated channel of the remained $\Delta{S}^l$ as $U_a$.
The target attribute-specific control units consist of $S^{l}_{U_a}$ and $F^{l}_{U_a}$.

The spatial distribution of the modification to $F^{l+1}$ caused by $\Delta{S}^l$ is determined by the spatial distribution of $F^l_{U_a}$.
But the activated region of $F^l_{U_a}$ is not always precisely aligned with the target region $r$, as shown in Fig.~\ref{fig:fig1}(b).
We thus manipulate $F^l_{U_a}$ to adjust the spatial distribution of $F^l_{U_a}$ and improve the effect of the change in $S^{l}_{U_a}$.
We trace back to the modulation styles $S^{l-1}$ and input feature $F^{l-1}$ of the previous layer that was used to compute to the feature maps $F^l_{U_a}$.
Instead of seeking a direction vector for $S^{l-1}$ and continue to track the previous layer that can modify $F^{l-1}$, we directly optimize to obtain a proper $\hat{S}^{l-1}$ as will be described in Sec.~\ref{sec:replacement}.

Fig.~\ref{fig:method} shows the overall flow of our manipulation pipeline.
Our modification consists of a optimized styles $\hat{S}^{l-1}$ and a direction vector $\Delta{S}^l$.
A few channels of $F^l$ are replaced by $F^{l}_{U_a}$ computed with $\hat{S}^{l-1}$, while other channels of $F^l$ keep untouched.
The original modulation style $S^l$ and $\Delta{S}^l$ are summed to form the new modulation style.
We will explain in detail how to obtain $U_r$, $U_a$, $\Delta{S}^l$ and $\hat{S}^{l-1}$ in the following sections.

\subsection{Detecting Feature Maps specialized to Synthesize Target Region}\label{sec:detect}

Feature maps that activate consistently in the semantic region $r$ across various generated images can be detected by analyzing the overlap of the high activated location with the semantic segmentation of the region $r$~\cite{bau2019gandissect}.
But we observe that this method often fails on small semantic regions(e.g., eyes) or low-resolution features in the generator.
We instead propose a more intuitive and straightforward way to determine where the high activated part of feature maps is located.

If the value of the region $r$ in $F_{k}^l$ is close to 0, then the gradient of $s^l_k$ that passed by the region $r$ in the gradient map of $F^{l+1}$ is also close to 0.
We then set the binary mask $M_r$ for the specific region $r$ as the gradient maps of the generated images, then apply back-propagation to compute the gradient with respect to each component of $S$.
For each semantic region $r$ in generated image and each channel index $i$ of $S$, we record the absolute value of the gradient, then normalize it use the sum of the values in $M_r$.
After repeating this process on 10K different images, we average the recorded values.
Let $g_r(s^l_k)$ denote the average of recorded values for $s_k^l$ and the semantic region $r$,
the set of indices of specialized feature maps for the region $r$ can be found as:
\begin{equation}\label{eq:correlation}
    U^l_r=\{k | \frac{g_r(s^l_k)}{\sum{g_c(s^l_k)}} > t^l_r, 0<{k}\le{n^l}\}
\end{equation}
where $c$ denotes each semantic region in generated images.
In this way, we can detect these channels with high activation in $r$ for most of the sampled images.

\begin{figure*}[t]
\begin{center}
\includegraphics[width=1.0\linewidth]{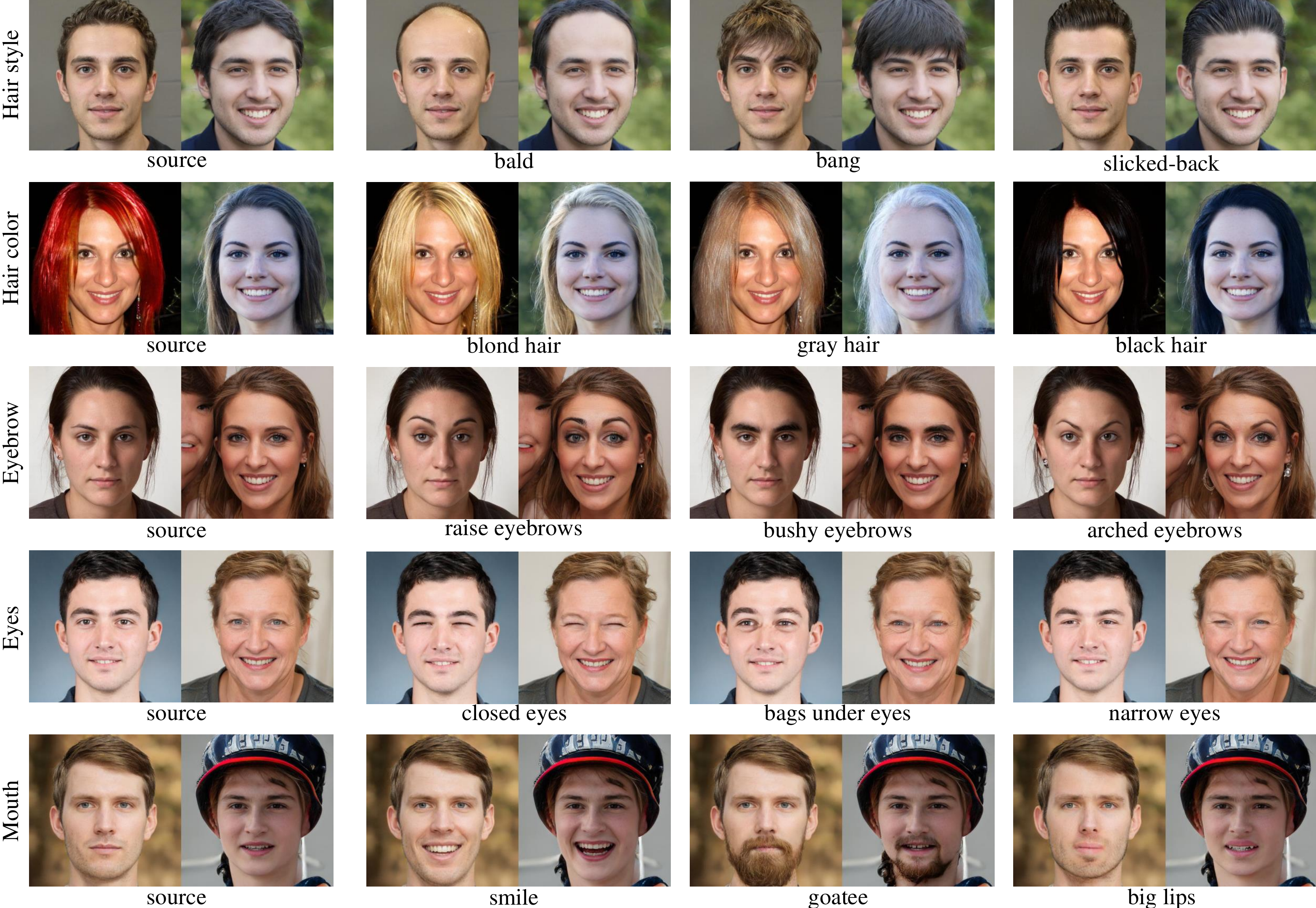}
\end{center}
\caption{
Examples of manipulating different attributes of different semantic regions.
}
\label{fig:results}
\end{figure*}

\subsection{Detecting Attribute-Specific Direction Vector}\label{sec:movement}

Different layers in StyleGAN2 usually control hierarchical attributes of region $r$.
Manipulating a specific attribute of local region $r$ usually requires only modifying modulation style for one single layer.
Previous works~\cite{shen2020interfacegan, harkonen2020ganspace,shen2021closedform} can identify a semantically meaningful direction vector in the $\mathcal{W}$ or $\mathcal{Z}$ space that can manipulates target attribute of $r$.
We first transformed these direction vectors found by other methods into the $\mathcal{S}$ space.
The attribute-specific direction vector $\Delta{S^l}$ can be obtained by zeroing all channels of the direction vector in the $\mathcal{S}$ space except for these channels in $U^{l}_r$.
The requirement for the initial direction vector is simple: it can correctly manipulate the local target attribute regardless of how much entanglement it introduces at the same time.
For these local attributes manipulation directions that cannot be found by the previous methods, we directly use the difference between the latent vector of one positive sample and one negative sample.
For example, we seek a face image $x_1$ with raised eyebrows and a face image $x_2$ without raised eyebrows.
We calculate the difference $\Delta{S}$ between the latent code $S_1\in \mathcal{S}$ and $S_2\in \mathcal{S}$ that used to synthesis image $x_1$ and $x_2$ respectively.
$\Delta{S}$ meets the requirement for a direction vector for raising the eyebrow.
In fact, by varying the target region $r$ and the layer $l$, we can construct directions corresponding to the differences in different attributes of different regions of image $x_1$ and $x_2$.

\subsection{Manipulate Attribute-Specific Control Feature maps}\label{sec:replacement}

Our goal is to manipulate $F^{l}_{U_a}$ to ensure the activated region of $F^{l}_{U_a}$  precisely aligns with the target region $r$, which means the pixels in the $r$ are entirely activated, and the values of pixels outside the region $r$ are close to zero.
$F^{l}_{U_a}$ is transformed from the input feature $F^{l-1}$ and modulation styles $S^{l-1}$ of the ${l-1}^{th}$ layer.
$F^{l-1}$ corresponding to different synthesized images share similar characteristics.
Therefore we can directly optimize to obtain a identical modulation style vector $\Tilde{S}^{l-1}$ for all images.
${\Tilde{S}^{l-1}}$ can be obtained by minimizing the following loss:
\begin{equation}\label{eq:loss}
    \mathcal{L} =  -\|F^{l}\odot{M_r^{l}} \| + \|F^{l} \odot{(1-M_r^{l})} \|
\end{equation}
\begin{equation}
    F^{l}=\Phi^{l-1}_{U_r}(\Tilde{S}^{l-1}, F^{l-1})
\end{equation}
where $\|\cdot\|$ denotes L1-norm, $\Phi^{l-1}$ denotes the whole the ${l-1}^{th}$ layer and $\odot$ denotes the Hadamard product.
$M_r^{l}$ is the binary mask, which is downsampled from the original mask $M_r$ to the same spatial resolution as $F^{l}$. 
However, directly replacing the original modulation style $S^{l-1}$ with $\Tilde{S}^{l-1}$ will leads too much change in $F^{l}_{U_a}$. 
So We introduce a replace factor $\alpha$ and set $\hat{S}^{l-1}=\alpha{\Tilde{S}^{l-1}}+(1-\alpha)S^{l-1}$.
Channels of $F^{l}$ are controlled in groups by feeding two modulation styles $S^{l-1}$ and $\hat{S}^{l-1}$ as shown in Fig.~\ref{fig:method}.

\section{Experiment}

\subsection{Experimental Settings}

\begin{figure*}[t]
\begin{center}
\includegraphics[width=1.0\linewidth]{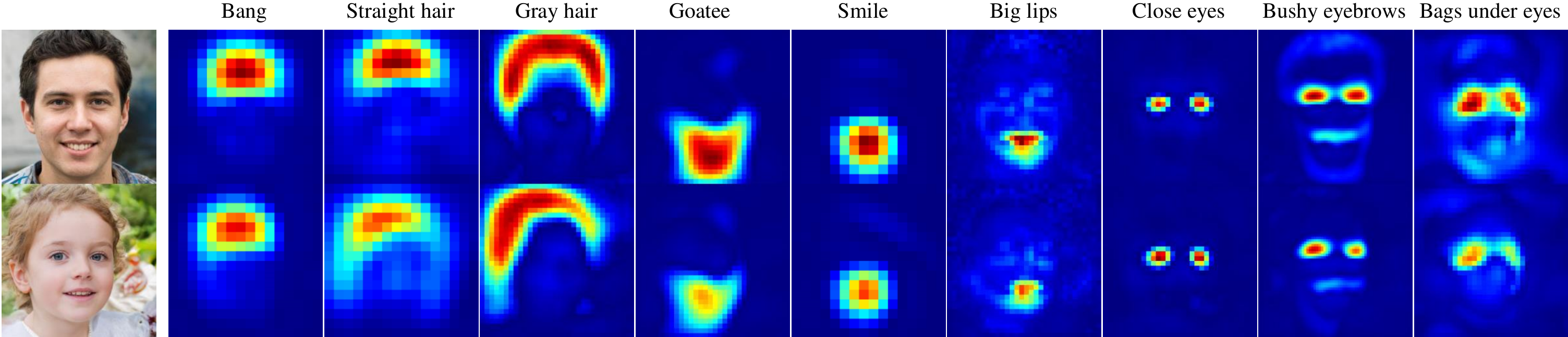}
\end{center}
\caption{
visualisation of attribute-specific control units for different local attributes.
Each channel of the manipulated feature $F^{l}$ is channel-wisely scaled by the direction vector $\Delta{S}^{l}$, then summed to get the final control units representation.
}
\label{fig:activation}
\end{figure*}

\begin{table*}[t]
\caption{Comparison of quantitative results measured by using different metrics. 'BE' and 'BH' are abbreviations for \textit{Bushy eyebrows} and \textit{Black hair} respectively.}
\label{table:quantitative_comparison} 
\centering
%&\phantom{a}
\resizebox{1.95\columnwidth}{!}{%
\begin{tabular} {lccc|ccc|ccc|ccc}
\toprule
      & \multicolumn{3}{c}{success rate ($\uparrow$)} &\multicolumn{3}{c}{region purity ($\uparrow$)} & \multicolumn{3}{c}{mean-AD ($\downarrow$)} & \multicolumn{3}{c}{FID ($\downarrow$)} \\
      \cmidrule{2-4}            \cmidrule{5-7}              \cmidrule{8-10} \cmidrule{11-13}
Method & Bangs & BE & BH  & Bangs & BE & BH & Bangs & BE & BH  & Bangs & BE & BH   \\
\midrule
InterFaceGAN~\cite{shen2020interfacegan}/$\mathcal{W}$  & 0.95 & \textbf{0.96} & \textbf{0.93} & 0.69 & 0.05 & 0.52 & 0.25 & 0.21 & 0.23 & 44.51 & \textbf{48.24} & \textbf{45.59} \\
InterFaceGAN~\cite{shen2020interfacegan}/$\mathcal{S}^l$  & 0.89 & 0.88 & 0.86 & 0.62 & 0.10 & 0.57 & 0.24 & 0.15 & 0.22 & 45.83 & 53.37 & 50.98\\
StyleSpace~\cite{wu2020stylespace}  & 0.78  & 0.84 & 0.58 & 0.77 & 0.31 & 0.78 & 0.25 & 0.11 & 0.16 & 46.29 & 53.35 & 56.26\\
Ours   & \textbf{0.97} & 0.93 & 0.84 & \textbf{0.91} & \textbf{0.58} & \textbf{0.86} & \textbf{0.20} & \textbf{0.10} & \textbf{0.14} & \textbf{44.35} & 48.92 & 49.80 \\
\bottomrule
\end{tabular}
}
\end{table*}

\begin{figure}[tb]
	\setlength{\tabcolsep}{1pt}	
	\begin{tabular}{cccccc}
		& {\footnotesize Original} & {\footnotesize ~\cite{shen2020interfacegan}/$\mathcal{W}$} & {\footnotesize ~\cite{shen2020interfacegan}/$\mathcal{S}^l$} &
		{\footnotesize ~\cite{wu2020stylespace}} & {\footnotesize Ours} \\
		
		\rotatebox{90}{\footnotesize \phantom{kk} Black Hair } &
		\includegraphics[scale=0.085]{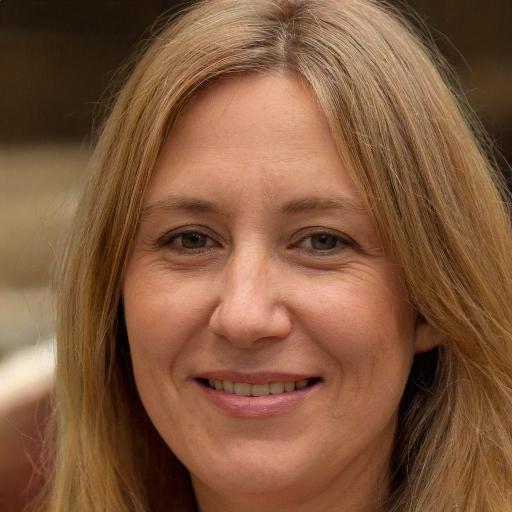} &
		\includegraphics[scale=0.085]{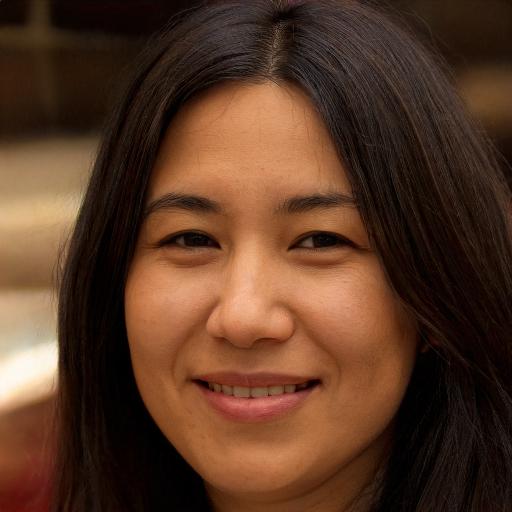} &
		\includegraphics[scale=0.085]{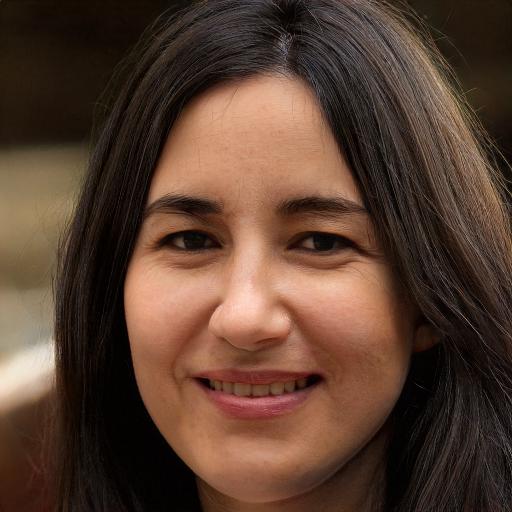} &
		\includegraphics[scale=0.085]{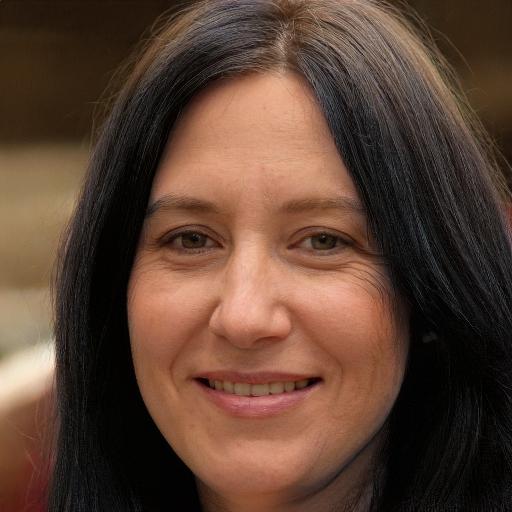} &
		\includegraphics[scale=0.085]{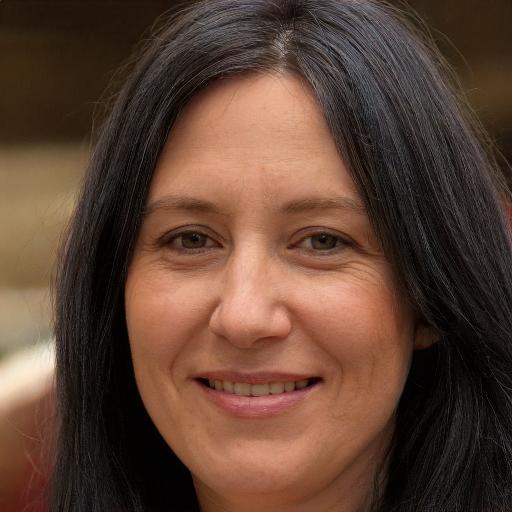}
		\\
		
		\rotatebox{90}{\footnotesize \phantom{kk}Gray Hair} &
		\includegraphics[scale=0.085]{./pics/compare/be_original.jpg} &
		\includegraphics[scale=0.085]{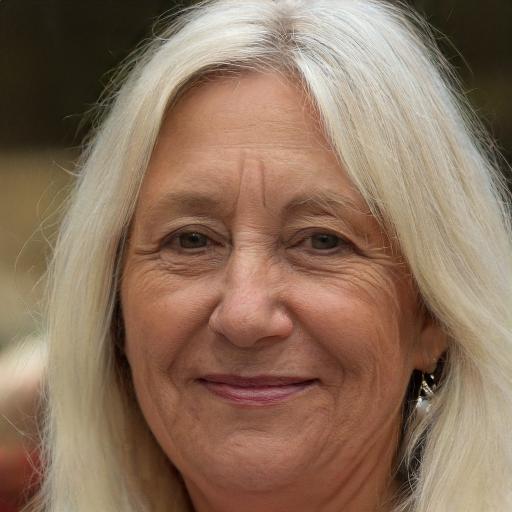} &
		\includegraphics[scale=0.085]{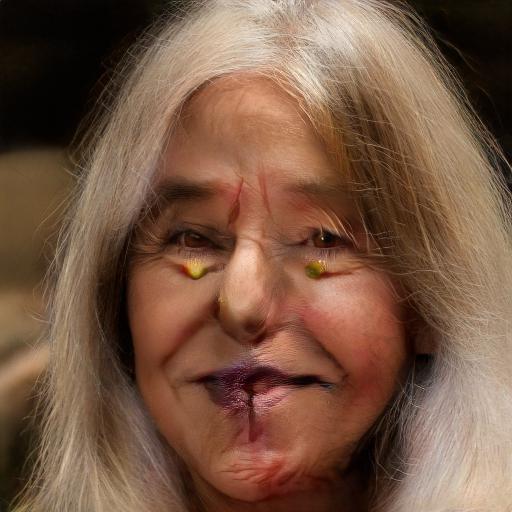} &
		\includegraphics[scale=0.085]{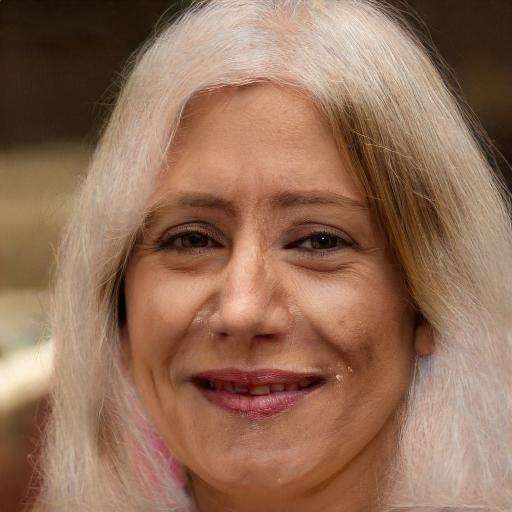} &
		\includegraphics[scale=0.085]{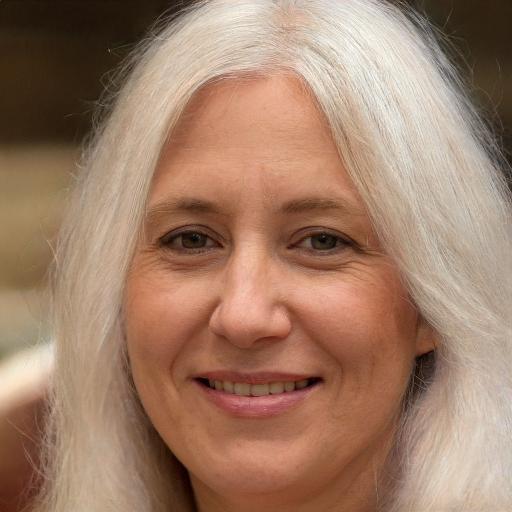}
		\\
		
		\rotatebox{90}{\footnotesize \phantom{kk} Eyebrows } &
		\includegraphics[scale=0.085]{./pics/compare/be_original.jpg} &
		\includegraphics[scale=0.085]{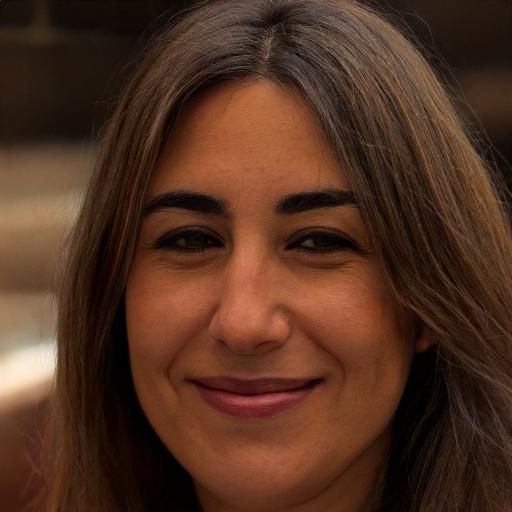} &
		\includegraphics[scale=0.085]{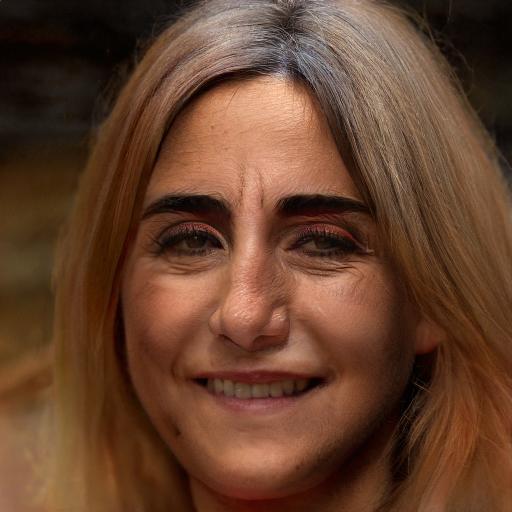} &
		\includegraphics[scale=0.085]{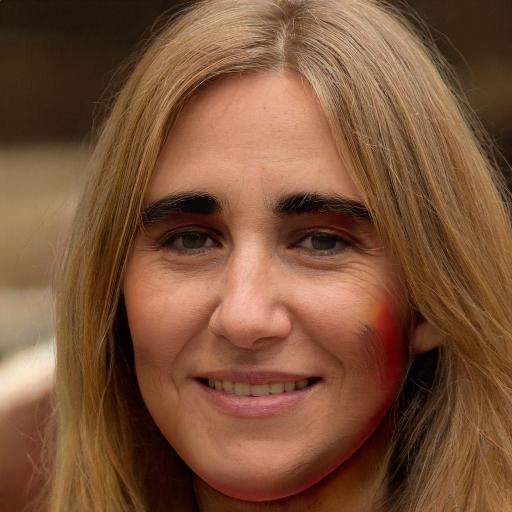} &
		\includegraphics[scale=0.085]{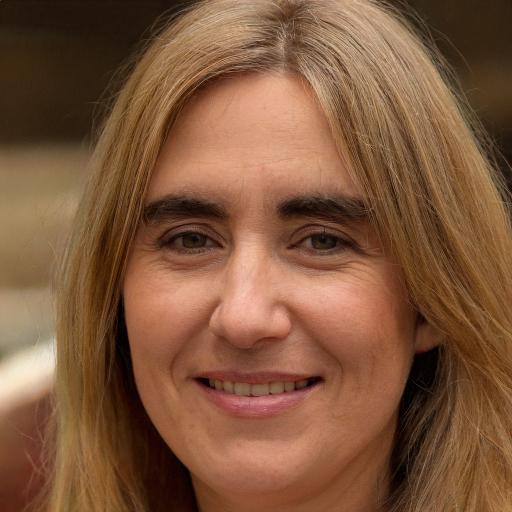}
		\\

% 		\rotatebox{90}{\footnotesize \phantom{kk} Goatee} &
% 		\includegraphics[scale=0.085]{./pics/compare/be_original.jpg} &
% 		\includegraphics[scale=0.085]{./pics/compare/goatee_ifg_w.jpg} &
% 		\includegraphics[scale=0.085]{./pics/compare/goatee_ifg_s.jpg} &
% 		\includegraphics[scale=0.085]{./pics/compare/goatee_stylespace.jpg} &
% 		\includegraphics[scale=0.085]{./pics/compare/goatee_ours.jpg}
% 		\\
		
		\rotatebox{90}{\footnotesize \phantom{kk} Bangs} &
		\includegraphics[scale=0.085]{./pics/compare/be_original.jpg} &
		\includegraphics[scale=0.085]{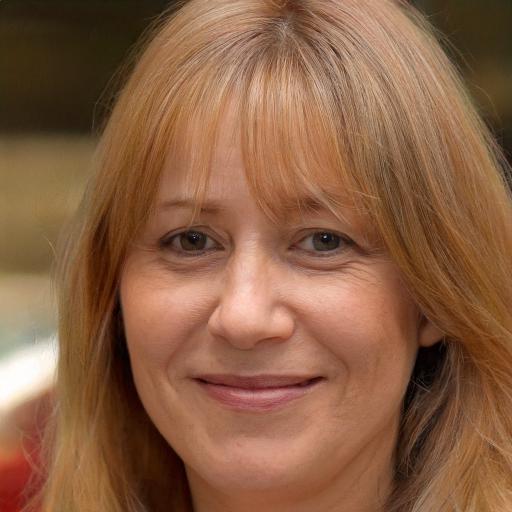} &
		\includegraphics[scale=0.085]{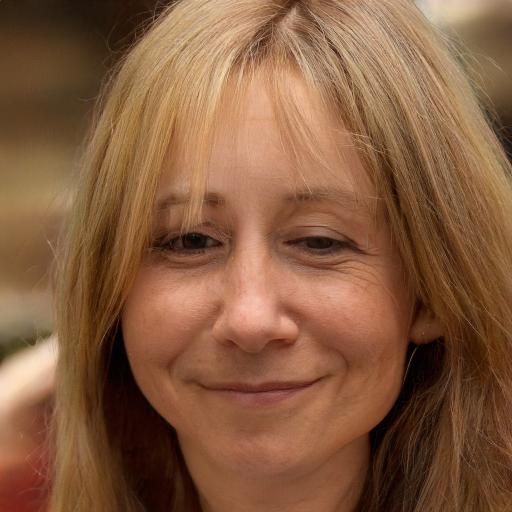} &
		\includegraphics[scale=0.085]{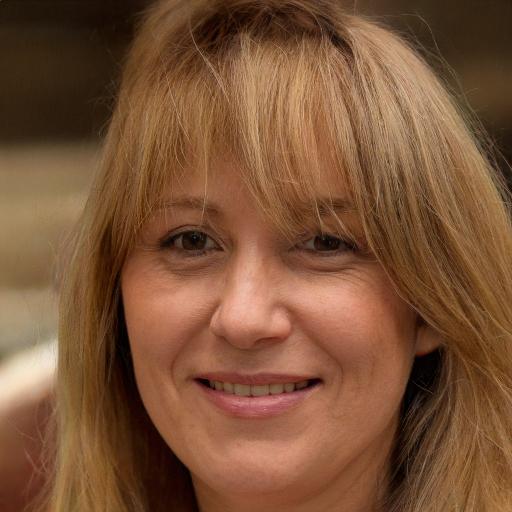} &
		\includegraphics[scale=0.085]{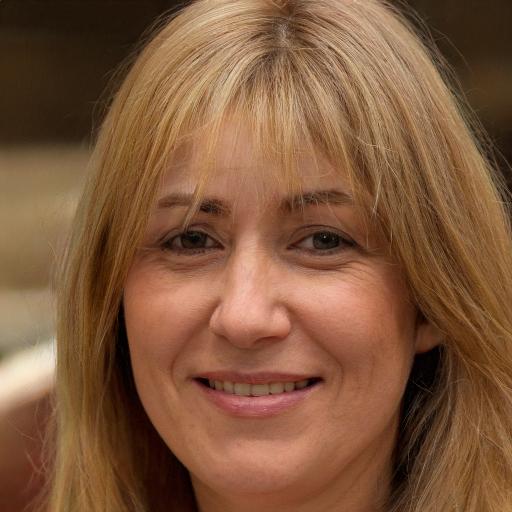}
		\\
	\end{tabular}
	
	\caption{Comparison with state-of-the-art methods InterfaceGAN~\cite{shen2020interfacegan} and StyleSpace~\cite{wu2020stylespace}.
		\vspace{-3mm}
	}
	\label{fig:comparison}
\end{figure}

We evaluate the proposed method on the face attribute editing task because of the easy access to the face component segmentation and attribute classifiers and the wide range of applications.
Our experiments are conducted on the StyleGAN2~\cite{karras2020analyzing} Generator pretrained on the Flickr-Faces-HQ Dataset (FFHQ)~\cite{karras2019style} with resolution $1024\times 1024$.
To obtain semantic maps for different facial components, we use the BiSeNet~\cite{yu2018bisenet} model pretrained on CelebAMask-HQ~\cite{CelebAMask-HQ} dataset.
3000 real face images from CelebAMask-HQ~\cite{CelebAMask-HQ} and 40 binary attributes annotations provided by \cite{liu2015faceattributes} are used to train a ResNet50~\cite{he2016deep} based multi-branch face attribute classifier for the following experiments.

We optimize $\hat{S}^{l-1}$ using stochastic gradient descent for each attribute editing task.
$\hat{S}^{l-1}$ is initialized with the average of styles fed into the ${l-1}^{th}$ layer.
We adopt Adam~\cite{kingma2014adam} optimizer with a learning rate of 0.1, batch size of 16 for 1K iterations to optimize $\hat{S}^{l-1}$.
We set empirically the threshold in Eq.~\ref{eq:correlation} between 0.1 and 0.2.
$\alpha$ is set as 0.5 for most attributes.
Details for the specific $l$ for various attributes and the comparison of different $\alpha$ can be found in \textbf{supplementary material}.

\subsection{Results of local attributes manipulation}

We validated our approach on a variety of local attribute editing tasks.
We apply InterfaceGAN~\cite{shen2020interfacegan} to find the initial direction vectors to be cropped corresponding to those attributes that have annotated by ~\cite{liu2015faceattributes}.
For other attributes that cannot train an available classifier, we directly use the difference between the modulation styles of a few annotated positive and negative samples as the direction vectors.
% %
% These semantic and spatial entanglement directions are then cropped as described by Sec.~\ref{sec:movement}, the feature maps in attribute-specific control units are replaced as described by Sec.~\ref{sec:replacement}.
% %
Figure~\ref{fig:results} plots the manipulation results on different attributes of different semantic regions in the human face.
We can see that our method can translate local attributes while keeping other regions in the image untouched.
It suggests that our method can achieve fine-grained controls on local semantic regions of generated images.
Note that, our proposed method can perform various local attribute editing tasks, which is much more than previous methods~\cite{shen2020interfacegan,wu2020stylespace,shen2021closedform,harkonen2020ganspace}.

Then we further visualize the control units for several attribute manipulations in Figure~\ref{fig:activation}. 
The top activated region of control units aligns with the target semantic region of different local attributes. 
It can be expected that the changes obtained by modifying these control units are spatial disentangled.

\begin{figure}[t]
\begin{center}
\includegraphics[width=1.0\linewidth]{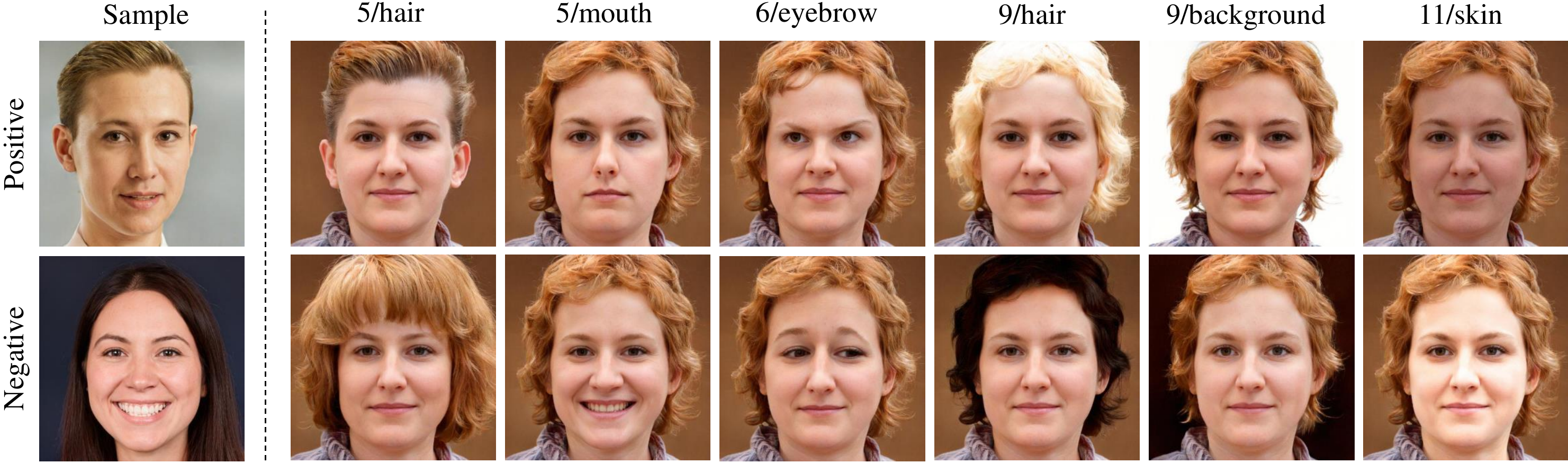}
\end{center}
\caption{Multiple local attribute editing directions obtained with only two samples. The layer index and the specified region is placed above the result images. 
}\label{fig:cut_off}
\end{figure}

\begin{figure*}[t]
\begin{center}
\includegraphics[width=1.0\linewidth]{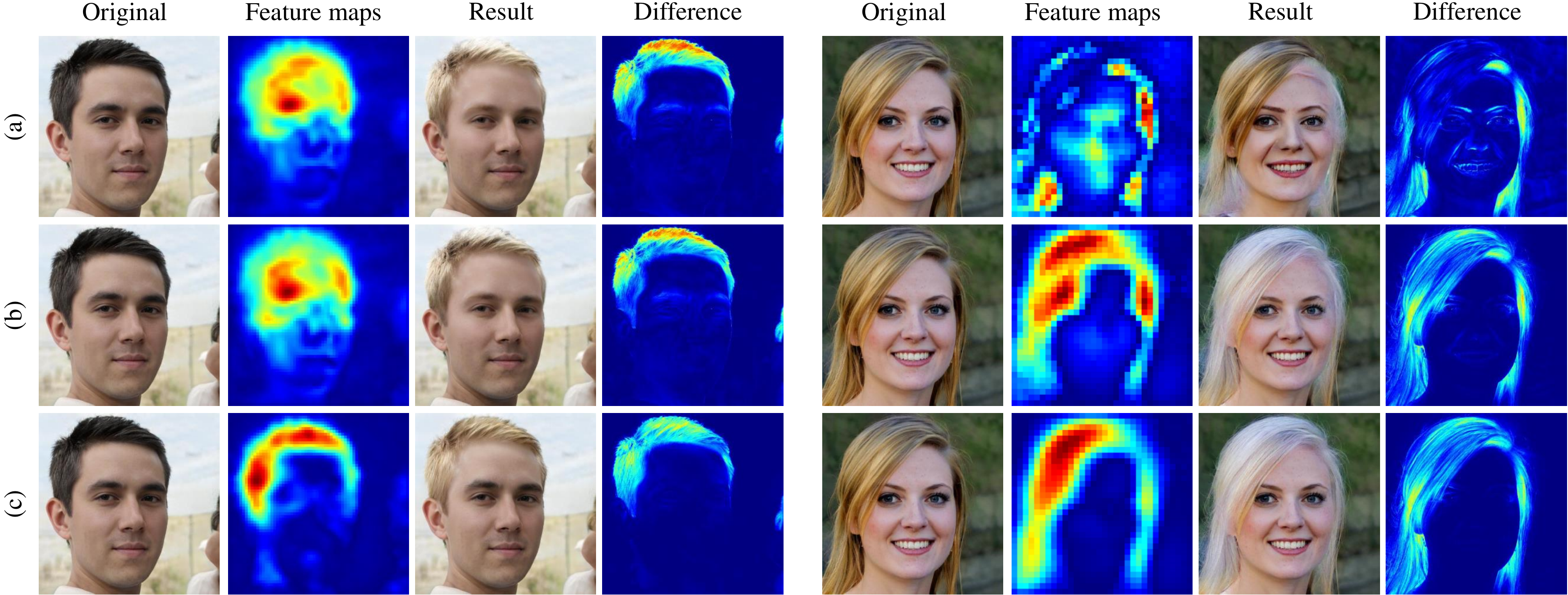}
\end{center}
\caption{
Comparison of different feature manipulation strategies for changing the hair color.
(a) No modification to the feature map.
(b) Moving the filter-wise styles for the control feature maps along the detected direction vector.
(c) Replacing these filter-wise styles with the optimized style vector.
}
\label{fig:replcement_importance}
\end{figure*}

\begin{figure*}[t]
\begin{center}
\includegraphics[width=1.0\linewidth]{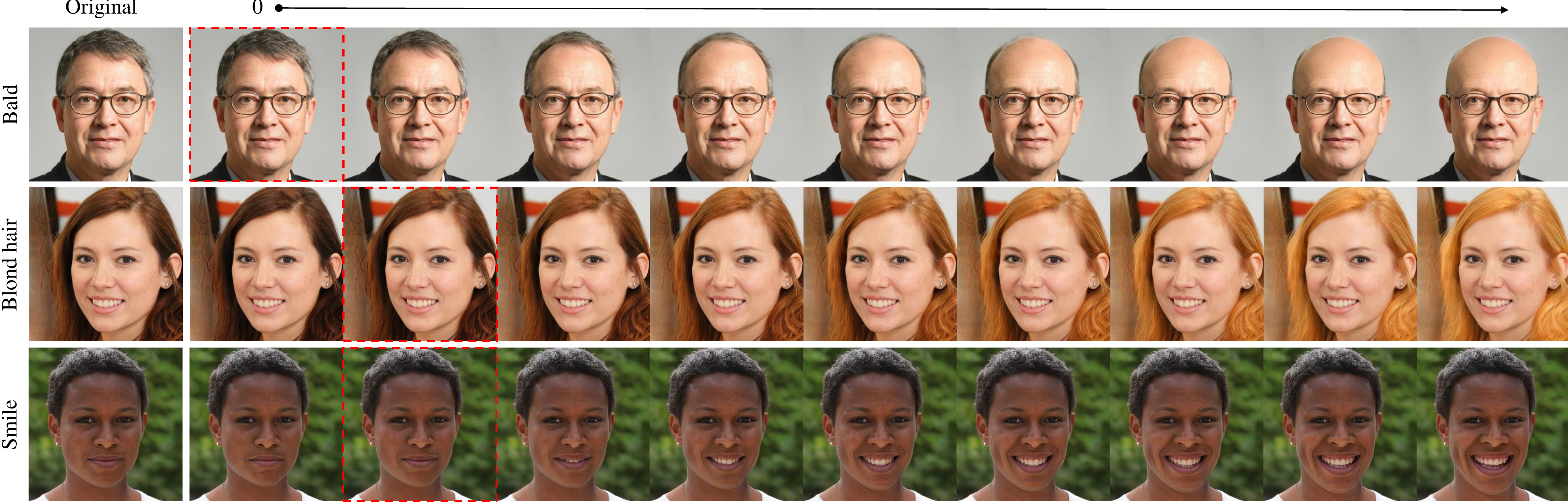}
\end{center}
\caption{
Visual results of continuous editing.
Each row shows the original image and the editing results obtained by increasing the move distance linearly from zero after replacing the critical feature maps.
The images bounded by red dashed lines indicate the closest editing result to the original images.
}
\label{fig:continuous_editing}
\end{figure*}

\begin{figure*}[t]
\begin{center}
\includegraphics[width=1.0\linewidth]{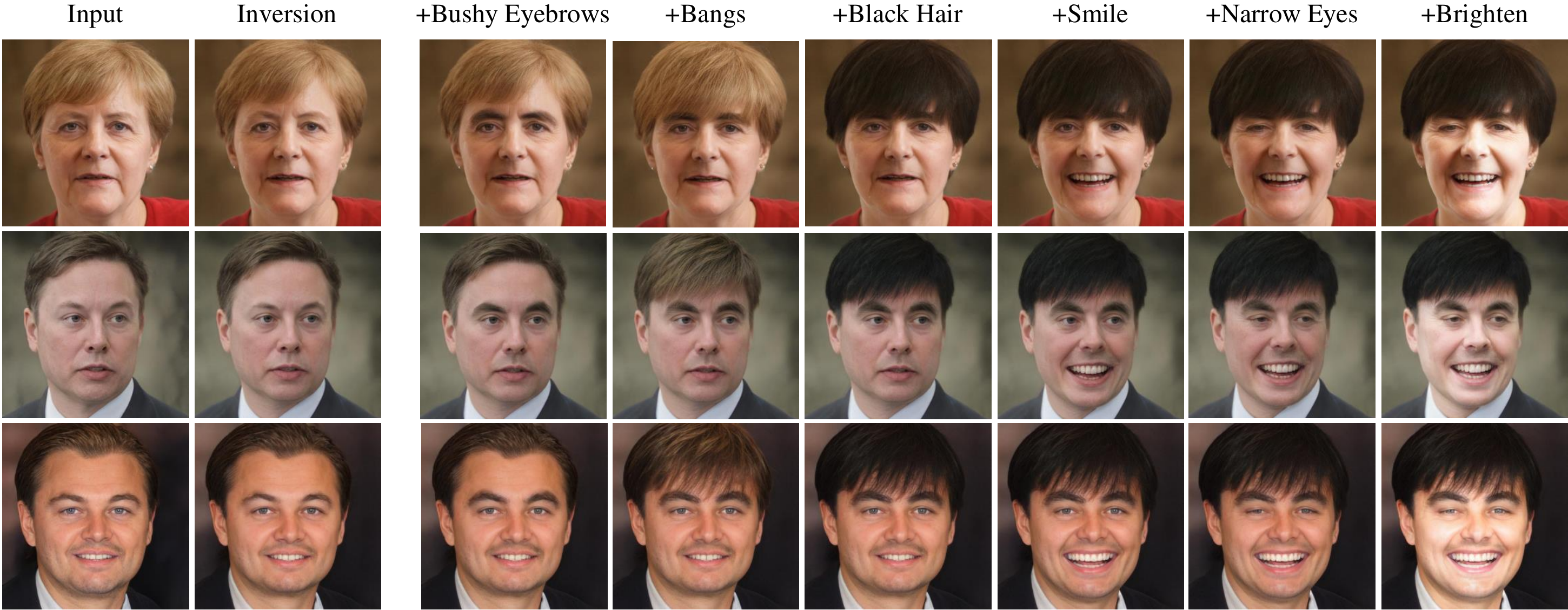}
\end{center}
\caption{
Manipulating real images using GAN inversion. 
The input images are first embedded in $W+$ space using the e4e encoder~\cite{tov2021designing}, then various facial attributes of the input are edited sequentially.
}
\label{fig:edit_real_img}
\end{figure*}

\subsection{Comparison with Previous Approach}

In this section we compare our approach with two state-of-the-art methods, i.e., InterFaceGAN~\cite{shen2020interfacegan} and StyleSpace~\cite{wu2020stylespace}. 
We implemented interfaceGAN~\cite{shen2020interfacegan} in the $\mathcal{W}$ space and the $\mathcal{S}^l$ space respectively.
The direction vectors in the $\mathcal{S}^l$ space only affect modulation style for the $l^{th}$ layer in StyleGAN2.
%
%
% There are many single-channel controls to manipulate \textit{Bang} provided by StyleSpace.
% %
We choose the top-1 channel sorted by StyleSpace in several single-channel controls that can manipulate the same attribute. 
For attributes \textit{Gray hair}, we use single-channel control found by StyleSpace~\cite{wu2020stylespace} as  $\Delta{S^l}$.
Other attributes use the cropped direction vectors in the $\mathcal{S}^l$ space.
We then replaced the original filter-wise styles used to compute the corresponding feature maps with the attribute-specific $\hat{S}^{l-1}$ for all attributes.
The step size for direction vector of different approaches is chosen such that it induces the same amount of change in the target local attribute.  
Figure~\ref{fig:comparison} show a qualitative comparison between different methods.
We can see that, compared with other methods, our method can precisely manipulate the given local attributes without affecting other attributes and regions. 
These results demonstrate that our proposed method can obtain both semantic and spatial disentangled controls.

% StyleSpace~\cite{wu2020stylespace} found various single channel controls for fine-grained local attributes manipulation.
% %
% These single channel controls naturally fit our requirements for $\Delta{S^l}$.
% %
% The main difference between our method and it is that StyleSpace does not modify the corresponding  feature maps when using a single channel control as $\Delta{S^l}$.
% %

We conduct quantitative and comprehensive experiments to further validate the superiority of our approach.
We followed the same setting in StyleSpace~\cite{wu2020stylespace}, 
sampled 3000 images without the target attribute and manipulated the sampled image to increase the logit value $l_t$ of the corresponding classifier.
$\Delta{l}_t$ is set as the standard deviation of the logit value for target attribute.
Not all images can increase the desired logit before they look less than realistic.
We measure the success rate of local translation.
We then adopt the Attribute Dependency(mean-AD~\cite{wu2020stylespace}) and Frechet Inception Distance (FID)~\cite{heusel2017gans} as metrics to measure the semantic disentanglement and fidelity of manipulated results.
FID is a commonly used metric to measure the diversity and quality of the generated samples.
Attribute Dependency measures the degree to which manipulation along a certain direction induces changes in logits of the classifier for other attributes.
To demonstrate the localization of the edits, we further introduce the region purity metric.
The region purity metric measures the proportion of Mean Squared Error (MSE) in the target semantic region among the MSE in the whole manipulated image.
Intuitively, spatial disentangled manipulations should induce smaller changes in other areas of the generated images.

According to results reported in Table.~\ref{table:quantitative_comparison}, using the direction vector in the $\mathcal{W}$ space to manipulate images allows traversing a more unrestricted distribution of generated images, providing visually plausible results.
In contrast, the direction in $\mathcal{W}$ may significantly change the unrelated region of the generated image. (eps. in the \textit{Black Hair} manipulation in Fig.~\ref{fig:comparison})
For example, the region purity of direction in $\mathcal{W}$ found by InterFaceGAN is only 0.05. 
Most of the modifications change the content of the unrelated regions which indicates the presence of severe spatial entanglement.
Restricting modifications to $S^l$ or a single channel of the $S^l$ can gradually increase the region purity metric and decrease mean-AD, i.e. reduce both spatial and semantic entanglement.
However, it also leads to a considerable drop in efficiency and fidelity at the same time.
Our method outperforms benchmark methods in terms of the metrics of the disentanglement, while our results are similar to the best method in terms of efficiency and image quality metrics at the same time.
This is because that $\hat{S}^{l-1}$ obtained with optimization increase the activation of the pixels in target region of $F^{U_a}$, the modulation style does not need to be moved very far to achieve the same logit changes. 

\subsection{Ablation Study}

\textbf{Results of feature maps grouping}~~We first select two synthesized images as positive and negative samples, respectively.
We calculated the difference between their modulation styles of different channel groups in different layers of StyleGAN2, following the approach described in Section~\ref{sec:detect}.
As shown in Fig.~\ref{fig:cut_off}, each of these sparse difference vectors corresponds to a semantic meaningful and spatial disentangled attribute manipulation.
This demonstrates that only channels in corresponding group is responsible for change the selected region of generated images, as stated in Eq.~\ref{eq:delta_F_2}.
Our proposed gradient-based strategy does accurately grouped the different channels of the intermediate features.

\textbf{The role of control feature maps manipulation.}~~We compare three strategies to demonstrate the importance of the replacement of the critical feature maps: 1) only moving the modulation style $S^l$ along the direction $\Delta{S^l}$. 2) modifying $S^l$ as previously described and moving the filter-wise styles used to compute $F^{l}_{U_a}$ along the direction $\Delta{S}^{l-1}$ identified with ~\cite{shen2020interfacegan}. 3) our full approach.
Fig.~\ref{fig:replcement_importance} compares the modified feature maps and the manipulation results of two hair color editing tasks.
After replacing filter-wise styles for $F^{l}_{U_a}$, the hair region in $F^{l}_{U_a}$ is entirely activated, and activation at other areas of $F^{l}_{U_a}$ can be ignored.
This allows the hair color to be completely translated without affecting the details like the eyebrow in the face area, as shown in Fig.~\ref{fig:replcement_importance}a and Fig.~\ref{fig:replcement_importance}c.
The second strategy failed in blond hair control and obtained suboptimal results for the blond hair attribute (Fig.~\ref{fig:replcement_importance}b).
We think this is because that the dataset biases dominate the direction vector found by ~\cite{shen2020interfacegan}.
\textit{Blonde hair} in the FFHQ dataset~\cite{karras2019style} is entangled with the race, but \textit{Gray hair} is uniformly present across different races.
Optimizing to obtain proper $S^{l-1}$ can overcome the dataset biases.

\textbf{Continuous Editing.}~~Our method introduces a nonlinear feature manipulation component.
We use the same optimized style to replace the ${l-1}^{th}$ layer's original filter-wise styles, regardless of how far $S^l$ walked.
We conducted experiments on several attribute editing tasks to observe the effect of this nonlinear module on continuous editing.
The results are shown in Figure~\ref{fig:continuous_editing}.
We can see that the synthesized image's target semantic region is changed continuously as the moving distance increases linearly.
Then we calculated and compared MSE error of each image in the interpolation process with the original image.
We found that the result without modifying the modulation style may not be the closest image to the original image.
This demonstrates that modification to feature maps in control units leads to minor changes in the synthesized images and these minor changes can be overcome by moving modulation along the found direction.

\subsection{Manipulation of Real Images}

We verify that our local attribute editing method works for real images in this section.
We first invert the input images to the latent codes with the GAN inversion method for further editing.
More concretely, we use the e4e encoder~\cite{tov2021designing} to embed the given input image into the $\mathcal{W}+$ space, and then manipulate multiple target attributes in turn.
Fig~.\ref{fig:edit_real_img} show some manipulation results, where our method shows impressive performance.
Each attribute added affects only a single semantic region of the input image, and subsequent attribute manipulations afterward do not affect the previous attribute manipulations.
These results demonstrate that the attribute-specific control units we found for each local attribute also are generalizable for real image editing.

\section{Conclusion}

In this work, we have shown that various local attributes are controlled by the few channels of specific intermediate features and their corresponding modulation styles.
We then proposed a simple method to detect these attribute-specific control units.
Fine-grained StyleGAN controls can be achieved by manipulating the modulation styles and feature maps in attribute-specific control units simultaneously.
We plan to identify control units for global semantic attribute manipulations such as pose and age in future work.
Our approach may also inspire more exploitation of the hidden representations in GAN in the future.

\begin{acks}
This work is supported by National Natural Science Foundation of China (No. 61876210), Science and Technology Commission of Shanghai Municipality (No. 19511120800).
\end{acks}

%%\begin{acks}
%%To Robert, for the bagels and explaining CMYK and color spaces.
%%\end{acks}
\clearpage
%\bibliographystyle{ACM-Reference-Format}
%\balance
%\bibliography{egbib}

\balance

\bibliographystyle{ACM-Reference-Format}

% \appendix
% \input{sections/appendix}
\end{document}